\documentclass[conference]{IEEEtran}

\usepackage{cite}
\usepackage{amsmath,amssymb,amsfonts}
\usepackage{amsthm}
\newtheorem{definition}{Definition}
\theoremstyle{definition}
\usepackage{algorithmic}
\usepackage{graphicx}
\usepackage{textcomp}
\usepackage{xcolor}
\usepackage{subcaption}
\usepackage{multirow}
\usepackage{booktabs}
\newcommand{\ra}[1]{\renewcommand{\arraystretch}{#1}}
\def\BibTeX{{\rm B\kern-.05em{\sc i\kern-.025em b}\kern-.08em
    T\kern-.1667em\lower.7ex\hbox{E}\kern-.125emX}}

\begin{document}

\title{Systematic Training and Testing for Machine Learning Using Combinatorial Interaction Testing}

\author{\IEEEauthorblockN{Tyler Cody$^{a*}$, Erin Lanus$^b$, Daniel D. Doyle$^a$, Laura Freeman$^a$}
$^a$\textit{National Security Institute, Virginia Tech}, Arlington Virginia, USA \\
$^b$\textit{The MITRE Corporation}, McLean Virginia, USA \\
$^*$Corresponding Author: tcody@vt.edu \\
}

\maketitle

\begin{abstract}

This paper demonstrates the systematic use of combinatorial coverage for selecting and characterizing test and training sets for machine learning models. The presented work adapts combinatorial interaction testing, which has been successfully leveraged in identifying faults in software testing, to characterize data used in machine learning. The MNIST hand-written digits data is used to demonstrate that combinatorial coverage can be used to select test sets that stress machine learning model performance, to select training sets that lead to robust model performance, and to select data for fine-tuning models to new domains. Thus, the results posit combinatorial coverage as a holistic approach to training and testing for machine learning. In contrast to prior work which has focused on the use of coverage in regard to the internal of neural networks, this paper considers coverage over simple features derived from inputs and outputs. Thus, this paper addresses the case where the supplier of test and training sets for machine learning models does not have intellectual property rights to the models themselves. Finally, the paper addresses prior criticism of combinatorial coverage and provides a rebuttal which advocates the use of coverage metrics in machine learning applications.

\end{abstract}


\section{Introduction}

Combinatorial interaction testing has been widely used in software testing to understand code coverage, fault location, and to develop pseudo-exhaustive test strategies. Testing machine learning (ML), however, introduces new challenges. Namely, ML depends on learning from data and has non-deterministic faults.

When a system function is \emph{learned from data}, instead of, e.g., hard-coded or hand-engineered, it introduces questions regarding how much data is needed, and regarding what data is needed, e.g., what domain should the data be collected from? Are there enough distinguishing features for class separation? What data surrounds a class and what data is needed to further distinguish it? Moreover, it is unclear how we ensure test data is adequate. That is, do we use a single, held-out set of testing data, or a leave-$n$-out procedure, or $k$-fold cross-validation, or both? Those procedures may be progressively more statistically rigorous, but they only test performance in independent and identically distributed settings. In real-world settings, operational environments are expected to differ from training environments. To address robustness to such differences, adequate testing may require an amount of out-of-distribution testing, and adequate training may require an amount of out-of-distribution training.

A system function that has \emph{non-deterministic faults} means that its faults are not necessarily reproducible, even with exhaustive testing. And to the extent that they are reproducible, the stochastic nature of outputs given by many ML models means that testing unique combinations one time may be inadequate. Moreover, outputs are dependent on the domain of the inputs, and the domains during testing and operation may differ from each other and from that seen in training. This non-stationarity challenges traditional combinatorial interaction testing, which typically focuses on comparing interactions seen in testing to an exhaustive universe of possible interactions \cite{kuhn2004software}.

Combinatorial coverage has been extended to help address these challenges \cite{lanus2021combinatorial}. However, Li et al. criticize recent applications of coverage to deep learning \cite{pei2017deepxplore, ma2018deepgauge, ma2019deepct} for its weakness to adversarial inputs, claiming that discretization creates a space where natural inputs are sparse and adversarial inputs are abundant \cite{li2019structural}. Furthermore, Li et al. claim that coverage is not correlated to test error on natural inputs. The results presented herein provide falsifying evidence of the latter claim, namely, that there is a correlation between test error on natural inputs and coverage.

This paper contributes a study of new systematic approaches to training and testing offered by combinatorial interaction testing using deep learning and the MNIST handwritten digit data set \cite{lecun1998mnist}. This paper demonstrates new approaches for using combinatorial interaction testing in machine learning for (1) constructing test sets, (2) constructing training sets, and (3) directing labeling efforts. While previous works have demonstrated specific uses of combinatorial coverage in deep learning \cite{pei2017deepxplore, ma2018deepgauge, ma2019deepct}, this paper demonstrates its use in training and testing broadly---agnostic to the particular machine learning solution method. Thus, this paper posits combinatorial interaction testing as a holistic approach to training and testing in ML, spanning white-box and black-box. Additionally, the presented results clearly refute criticisms in the literature that coverage is not correlated to test error on natural inputs.

This paper is structured as follows. First, a background and preliminaries are given on combinatorial interaction testing and on training and testing in machine learning. Then, methods using combinatorial coverage to address the identified concerns are presented using deep learning and MNIST. Subsequently, the presented results are used as the basis for a discussion of criticisms of combinatorial interaction testing in the training and testing of ML. Lastly, the paper closes with a synopsis and conclusions.

\section{Background}

\subsection{Training and Testing in Machine Learning}

ML algorithms need data from which to learn concepts, features, and hidden patterns in order to develop models for prediction.  This training data must be collected and/or synthetically generated, and it should represent that which the trained ML model will encounter in testing and in the real world. To achieve this, training data is often randomized, supplemented, and/or augmented to fill gaps and increase prediction performance \cite{shorten2019survey, anderson2014synthetic, dahmen2019synsys, raghunathan2021synthetic}.

Despite the complexity of constructing sets of training data, testing ML models often takes on a singular focus of how accurate is a given model in predicting the class or value of a specific item, object, etc. In the extreme, this singular focus considers rigour in testing to be achieved by training and evaluating on different but identically distributed subsets of data, as in k-fold cross-validation and leave-one-out stability \cite{arlot2010survey, vehtari2017practical}.

However, there are multiple factors to consider when testing \cite{murphy2007approach, goodfellow2017challenge, zhang2020machine}. What is the accuracy of the model given training, validation, and test data? Are the training and test data accurate, i.e., are inputs measured correctly and outputs labeled correctly? Literature on testing in ML is concerned with these and related questions, and with their great variety of answers. In a survey of the ML testing literature, an initial search found 1305 related papers that were subsequently down-selected to a relevant 37 publications between 2007 to 2019 \cite{braiek2020testing}.

Braiek and Khomh consider testing as either black-box or white-box, where the latter indicates that test procedures have access to or full-view of ML models' inner-workings and the former indicates that test procedures only have access to the input-output functionality of ML models \cite{braiek2020testing}. Common black-box methods include generative models, which leverage probability distributions fit to data, and adversarial ML, which considers the effect of alterations to inputs on performance.

White-box methods are numerous and varied. Common white-box methods related to this paper include metamorphic testing. Metamorphic testing uses structural properties of the relation between the inputs and outputs of a ML model to detect incorrect outputs without labels \cite{xie2011testing}. Combinatorial coverage also has a focus on structure.

There are various uses of coverage in white-box testing. In deep learning, recent work considers notions of coverage within layers of deep neural networks \cite{pei2017deepxplore, ma2018deepgauge, ma2019deepct}. Whereas methods such as DeepGauge \cite{ma2018deepgauge} and DeepCT \cite{ma2019deepct} consider coverage with respect to the internal \emph{layers} of neural networks, sometimes termed the neuron- or layer-levels, the presented work considers coverage over the latent space of an auto-encoder trained on the \emph{data} and its relation to the performance of a neural network classifier. Thus, in comparison to prior work involving combinatorial interaction testing and deep learning, the methodology presented herein is black-box with respect to the classifier. This is advantageous, as many times classifiers are proprietary or otherwise have inaccessible inner-workings. Consider, customers of ML services, e.g., governments, acquire ML models without acquiring full intellectual property rights to those models, and so cannot access algorithms. Also, in contrast to prior work, the work presented herein uses a variant of coverage that compares relative coverage between testing and training data \cite{lanus2021combinatorial}, as will be explained in the following.

\subsection{Combinatorial Interaction Testing}

Combinatorial interaction testing concerns the empirical testing of the interfaces between components of systems. It has been empirically shown that, in software, a limited number of interactions among components are responsible for nearly all failures \cite{kuhn2004software}. This finding supports the claim that high levels of assurance in component integration does not require exhaustive testing, and, thus, empirically supports the use of combinatorial approaches \cite{nie2011survey}.

In ML, combinatorial interaction testing offers an approach to testing that can scale to large input spaces and capture phenomena that emerge at the integration and systems levels, thereby addressing key challenges to testing in machine learning \cite{marijan2019challenges}. It has been used for testing machine learning components in autonomous driving \cite{tuncali2018simulation}, studying hidden layers in deep learning \cite{ma2019deepct}, and for explainability of machine learning \cite{kuhn2020combinatorial}. MNIST has been widely used in studies of combinatorial coverage for deep learning \cite{pei2017deepxplore, ma2018deepgauge, ma2019deepct, li2019structural}. 

In contrast to the traditional use of probability theory in machine learning, combinatorial interaction testing uses set theory. In testing and evaluation for machine learning, this results in a focus on set differences instead of distances between probability distributions.

In combinatorial interaction testing for ML, \emph{factors} are the dimensions of the input space $\mathcal{X}$ and output space $\mathcal{Y}$. For example, an unmanned aerial vehicle may have factors from sensor readings in $\mathcal{X}$ such as ground speed, radar, and GPS. The \emph{values} of those factors are their events, e.g., `5 meters per second' is a value of the ground speed factor. In combinatorial interaction testing, continuous-valued factors must be discretized to a finite set of values.

A $t$-way \emph{value combination} is a $t$-tuple of (factor, value) pairs. If there are $k$ factors in $\mathcal{X} \times \mathcal{Y}$, then each element $(x, y) \in \mathcal{X} \times \mathcal{Y}$ contains $k \choose t$ $t$-way value combinations. A 3-way value combination in an unmanned aerial vehicle could be a specific combination of values for ground speed, radar, and GPS, e.g., `5 meters per second' with `no obstruction within 10 meters' in `region A'. A 2-way value combination could be a specific combination of values for radar and GPS, e.g., `no obstruction within 10 meters' in `region A'.

\emph{Combinatorial coverage}, also referred to as total $t$-way coverage, is a metric for the proportion of valid $t$-way value combinations that appear in a set \cite{kuhn2013combinatorial}. Value combinations that appear are considered \emph{covered}. Those that do not are considered \emph{not covered}. Combinatorial coverage is defined formally as follows.

\begin{definition}{\emph{$t$-way Combinatorial Coverage.}} \\
    Consider a universe with $k$ factors such that $\mathcal{U}$ is the set of all valid $k$-way value combinations. Let $\mathcal{U}^t$ be the set of valid $t$-way combinations. Given a set of data $D \subseteq \mathcal{U}$, let $D^t$ define the set of $t$-way value combinations appearing in $D$. The $t$-way combinatorial coverage of $D$ is
    $$CC^t(D) = \frac{|D^t|}{|\mathcal{U}^t|},$$
    where $|D|$ denotes the cardinality of $D$. 
\end{definition}

In comparing training and test sets of data, we cannot necessarily treat them as $D^t$ and $\mathcal{U}^t$ because training and test data may not be subsets of each other. A metric of coverage developed explicitly for ML, \emph{set difference combinatorial coverage} (SDCC), is alternatively tasked with comparing value combinations seen in training, to those seen in testing, to those seen in operation---all relative to each other, not relative to all possible value combinations. SDCC extends combinatorial coverage with set difference as follows \cite{lanus2021combinatorial}.

\begin{definition}{\emph{$t$-way Set Difference Combinatorial Coverage.}} \\
    Let $D_S$ and $D_T$ be sets of data, and $D{_S}^t$ and $D{_T}^t$ be the corresponding $t$-way sets of data. The set difference $D{_T}^t \setminus D{_S}^t$ gives the value combinations that are in $D{_T}^t$ but that are not in $D{_S}^t$. The $t$-way set difference combinatorial coverage is
    $$SDCC^t(D_T, D_S) = \frac{|D{_T}^t \setminus D{_S}^t|}{|D{_T}^t|}.$$
\end{definition}
\noindent Restated, it is the proportion of $t$-way value combinations appearing in $D_T$ but not $D_S$. $SDCC^t$ is bounded $[0, 1]$ where $1$ indicates no overlap, i.e., $D{_T}^t \cap D{_S}^t = \emptyset$, and $0$ indicates that $D{_T}^t \subseteq D{_S}^t$. In other words, if $D_T$ is the testing data and $D_S$ is the training data, $0$ indicates that all testing combinations are present in the training data and $1$ indicates that none are present. Note, we use $t=2$ throughout the paper and denote $SDCC^2$ as $SDCC$. 

\section{Methods}

In the following, correlations between accuracy and SDCC are demonstrated. Then, SDCC's use for constructing test sets, constructing training sets, and directing labeling efforts is investigated.

\subsection{SDCC on MNIST}

\begin{figure}[t]
    \centering
    \includegraphics[width=0.50\textwidth]{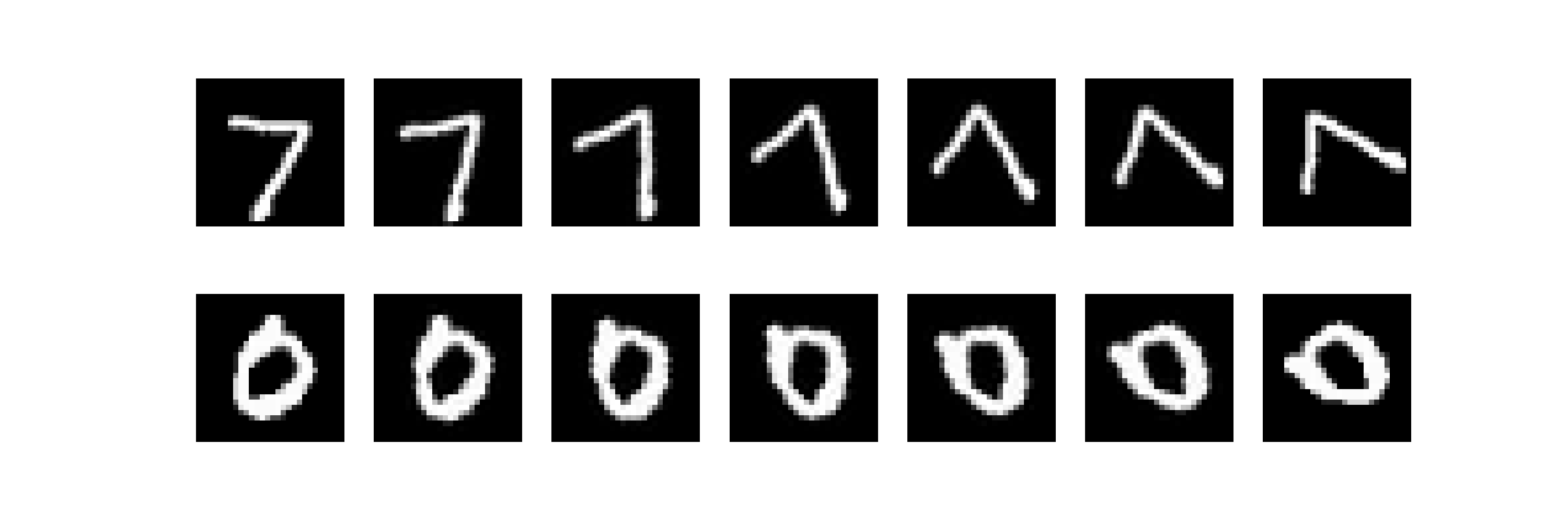}
    \caption{Example digits rotated 0, 15, 30, 45, 60, 75, and 90 degrees from left to right.}
    \label{fig:digits}
\end{figure}

SDCC is demonstrated on the MNIST handwritten digits data \cite{lecun1998mnist}. The data consists of 28$\times$28 pixel grayscale images of digits 0-9 and has been canonical for research in combinatorial interaction testing for deep learning \cite{pei2017deepxplore, ma2018deepgauge, ma2019deepct, li2019structural}. Herein, 1k images of each digit serve as the basis for several studies. Images are rotated counter-clockwise by 15, 30, 45, 60, 75, and 90 degrees, without resizing, as shown in Figure \ref{fig:digits}. This creates 60k rotated images.

Instead of pixel values, four discrete factors are used in order to study SDCC on MNIST. These factors are simple, general, and tractable, and avoid issues with meaningless interactions in the space of pixel values. The factors are as follows.
\begin{itemize}
    \item \textbf{Digit.} The digit label is a discrete factor: 0, 1, ..., 8, 9.
    \item \textbf{Circle.} Images with digit 0, 6, 8, and 9 are assigned True for their closure and other images are assigned False.
    \item \textbf{Density.} Images are separated into four bins using the 0.25, 0.50, and 0.75 quantiles of their average pixel value.
    \item \textbf{AE Region.} Images are assigned values corresponding to partitioned regions in the latent space of an autoencoder (AE) \cite{baldi2012autoencoders}.
\end{itemize}
\noindent The circle and density factors are more or less invariant to rotation. So, in order to have a factor dependent on rotation, the AE region factor is introduced. Since $t=2$, each image has six 2-way value combinations. The choice of $t=2$ is made because in initial experiments, the digit and AE region factors were in most $3$-way and $4$-way combinations. The AE region factor merits further explanation. 

To create the AE region factor, we train an AE using 1k images of each non-rotated digit for 20 epochs. The AE consists first of a network which encodes images into a low-dimensional latent space, and second of a network that reconstructs images given samples from the low-dimensional latent space. Details on architecture can be found in the Appendix. 

\begin{figure}[t]
    \centering
    \includegraphics[width=0.50\textwidth]{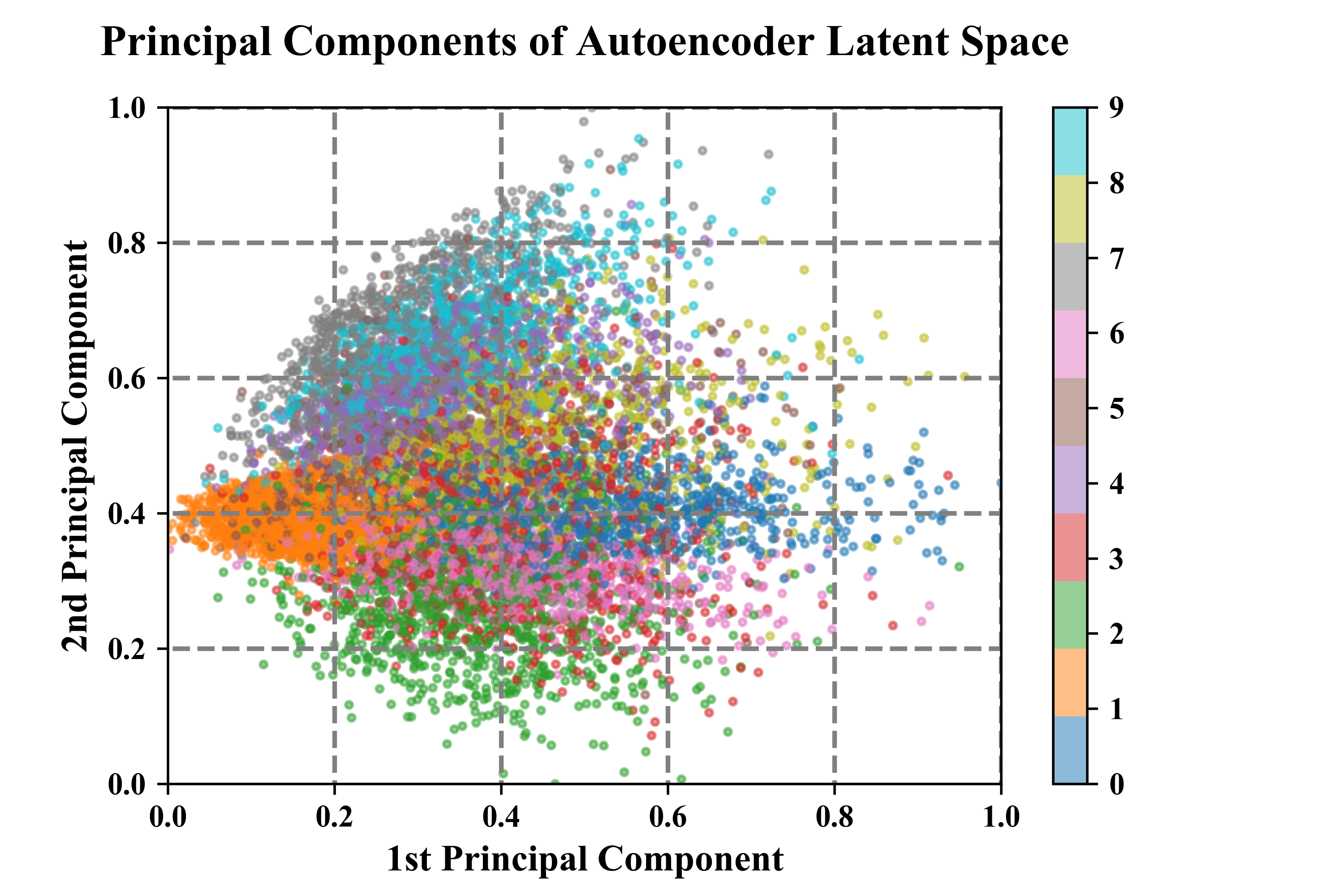}
    \caption{Principal components of AE's latent space, colored by digit, with region divisions shown by grey, dashed lines.}
    \label{fig:pca_ae}
\end{figure}

Discrete AE regions are created in two steps. First, images from all rotations are encoded in the AE's latent space, the encoded images are transformed into their first two principal components \cite{wold1987principal}, and the resulting data are scaled [0, 1]. Second, the [0, 1] $\times$ [0, 1] square is partitioned into 25 equal regions as shown in Figure \ref{fig:pca_ae}. 

Herein, convolutional neural networks (CNNs) are used to classify digits. Details on architecture can be found in the Appendix. In order to conduct a basic investigation of the correlation between SDCC and accuracy, a CNN termed CNN$_0$ is trained to classify non-rotated digits using 9k non-rotated images. Then, CNN$_0$ is evaluated on a test set of 1k non-rotated images and on 10k images associated with all other angles.

Additionally, SDCC(0, $\theta$) and SDCC($\theta$, 0) are computed for each angle $\theta$. Recall, the former describes the number of value combinations for non-rotated digits that do not appear in the rotated data, and the latter describes the number of value combinations for rotated digits that do not appear in the non-rotated data, based on the four defined factors. The results are shown in Table \ref{table:corr}.

\begin{table}\centering
\ra{1.3}
\begin{tabular}{@{}rrcrr@{}}\toprule
\multirow{2}{*}[-4pt]{{Angle $\theta$}} 
& \multirow{2}{*}[-4pt]{{Accuracy}}
& \phantom{abc} & \multicolumn{2}{c}{SDCC}\\ 
\cmidrule{4-5}
&&& (0, $\theta$) & ($\theta$, 0)\\ \midrule
0  & .99 &&   0 &   0\\
15 & .95 && .04 & .07\\
30 & .75 && .06 & .05\\
45 & .44 && .09 & .10\\
60 & .27 && .15 & .07\\
75 & .20 && .25 & .07\\
90 & .17 && .28 & .09\\
\bottomrule
\end{tabular}
\caption{Accuracy of CNN$_0$ on 1k held-out, non-rotated images and on 10k images rotated for each angle $\theta$, alongside corresponding SDCC values.}
\label{table:corr}
\end{table}

The SDCC(0, $\theta$) monotonically increases as the accuracy of CNN$_0$ decreases. In other words, when the test data covers less of the training data, the accuracy decreases. Note, initial experiments found that if the AE region factor is excluded, there is no correlation between coverage and error. This is because the circle and density factors are more or less invariant to rotation. As will be highlighted, the role of the AE region factor varies depending on the use of SDCC, e.g., for constructing test or training sets, or directing labeling. In the following subsections, the relationship between SDCC and accuracy is explored further by considering these various uses.


\subsection{Constructing Test Sets}

As noted, testing classifiers on data that is identically distributed to that which they are trained on is inadequate to assess real-world performance. Instead, before testing, SDCC can be used to partition data into samples whose images have covered value-combinations and samples whose images have at least one not covered value-combination. In essence, SDCC uses interactions between factors to identify the types of images that one would expect to be more difficult to classify because they are dissimilar in composition from those seen in training. This provides a kind of worst-case analysis for the classifier, offering a new mechanism for testing robustness to machine learning engineers. Instead of focusing on the stability of a classifier to samples of different composition drawn from the same distribution, SDCC-based testing focuses on separating samples into those with structurally representative and structurally challenging compositions.

To investigate the use of SDCC in constructing test sets, the accuracy of CNN$_0$ is evaluated on samples from each angle of rotation which (1) have value combinations that are covered by the non-rotated data\footnote{i.e., CNN$_0$'s training data} and which (2) have value combinations that are \emph{not} covered by the non-rotated data. The process of partitioning images as covered or not covered is as follows. For each angle $\theta$, SDCC($\theta$, 0) is used to identify those rotated images with value combinations that are not in the non-rotated data. Images have six 2-way value combinations, so any individual image may have many value combinations that are not in the non-rotated data. An image only needs one missing value combination to be considered not covered. All other images are considered covered.

CNN$_0$ is evaluated on samples of covered and not covered images. First, images are considered covered in the strict sense defined above. Second, the strict definition of covered is relaxed. The relaxed notion of covered considers an image as not covered if it belongs to an AE region that has a not covered image. The motivation behind this relaxation is to test if the strict focus on testing unseen value combinations is significantly different than constructing tests by sampling from regions in AE's latent space that have images with unseen value combinations. The results are shown in Table \ref{table:test}.

\begin{table}\centering
\ra{1.3}
\begin{tabular}{@{}rrrcrr@{}}\toprule
\multirow{2}{*}[-4pt]{{Angle $\theta$}}
& \multicolumn{2}{c}{Covered Images} & \phantom{abc} & \multicolumn{2}{c}{Not Covered Images}\\ 
\cmidrule{2-3} \cmidrule{5-6}
& \# Images & Accuracy && \# Images & Accuracy\\ \midrule
Strict\\
15 & 9839 & .96 && 161 & .81\\
30 & 9625 & .76 && 375 & .51\\
45 & 8317 & .49 && 1683 & .18\\
60 & 7874 & .31 && 2126 & .10\\
75 & 7543 & .24 && 2547 & .09\\
90 & 6848 & .21 && 3152 & .09\\
Relaxed\\
15 & 7127 & .96 && 2873 & .92\\
30 & 5415 & .80 && 4585 & .69\\
45 & 3172 & .55 && 6828 & .39\\
60 & 2155 & .31 && 7845 & .26\\
75 & 1975 & .17 && 8025 & .21\\
90 & 1394 & .14 && 8606 & .18\\
\bottomrule
\end{tabular}
\caption{Accuracy of CNN$_0$ on samples of images whose value-combinations are covered and not covered by CNN$_0$'s training sample. The top concerns images corresponding to the strict definition of covered from SDCC and the bottom concerns relaxed definition involving the AE's latent space.}
\label{table:test}
\end{table}

\begin{figure}[t]
    \centering
    \includegraphics[width=.50\textwidth]{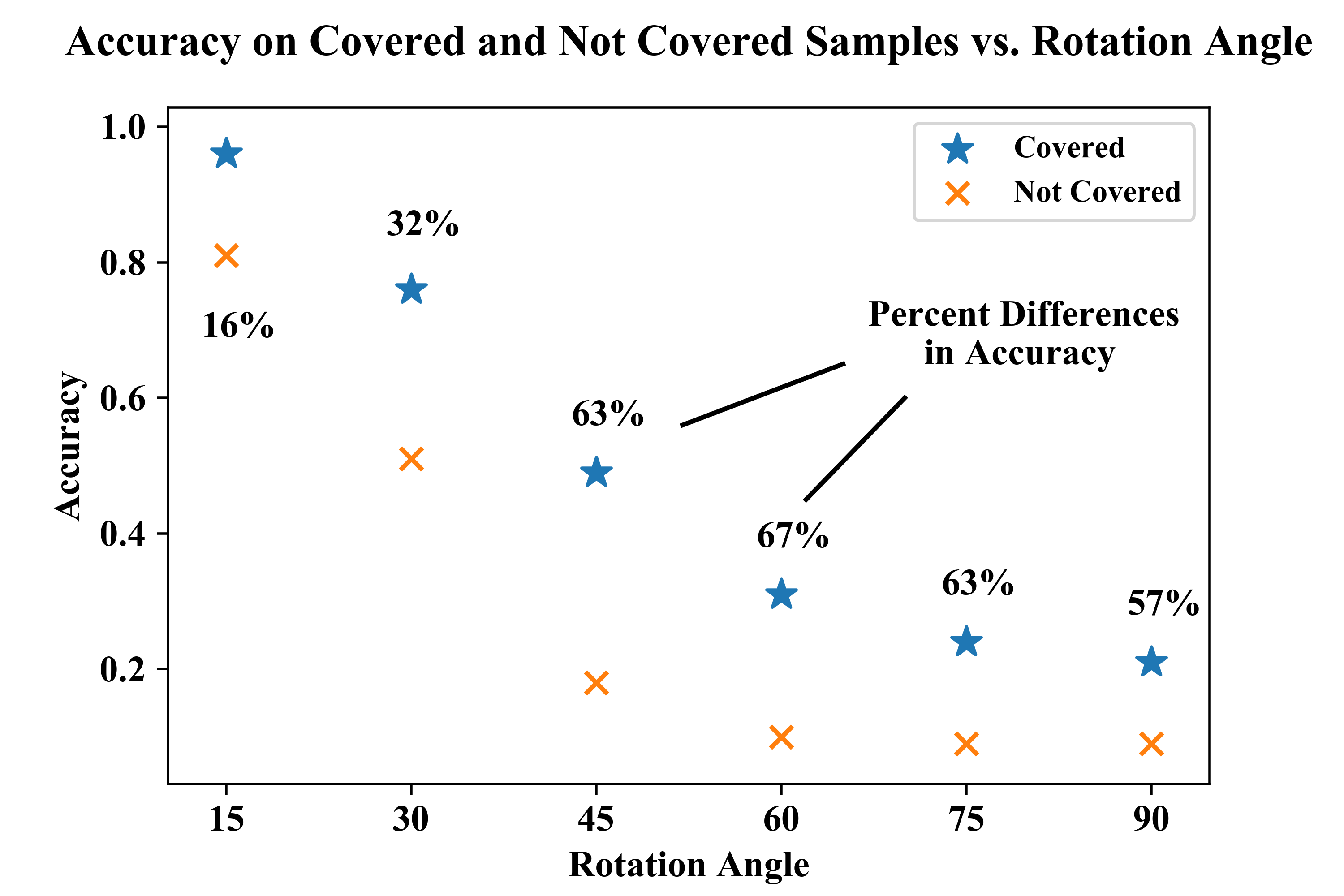}
    \caption{Accuracy difference on samples of covered and not covered images for each angle with percent differences.}
    \label{fig:test_set}
\end{figure}

There is a drastic difference in classification accuracy between samples of covered images and samples of not covered images. Figure \ref{fig:test_set} plots the two sets of accuracies for each angle, as well as denotes the percent difference between the two. For rotation angle 15, which had a relatively high performance on the randomly sampled test set as shown in Table \ref{table:corr}, the accuracy was only 16\% different. But across the other angles, the maximum difference was 67\% and the minimum was 32\%. This confirms the assertion that SDCC can be used to identify challenging test sets.

Furthermore, in comparing the strict and relaxed definitions of covered, the results show that constructing test sets by sampling from AE regions which have not covered images does not create nearly as challenging a test as sampling images with not covered value combinations directly. This illustrates the value of the interaction in defining set differences rather than univariate summaries of covered versus not covered. The bottom half of Table \ref{table:test} shows that, under the relaxed definition, CNN$_0$ has similar accuracy on covered and not covered images. Thus, while the AE region factor is needed to correlate SDCC with accuracy, the other factors are needed to add the necessary detail to identify difficult-to-classify edge cases.

\subsection{Constructing Training Sets}

Because of the unknowns associated with real-world deployment, defining notions of adequacy for training sets is difficult. Clearly, if the distributions of inputs or outputs are expected to change, then adequacy does not equate to a percentage size of held-out training data or a number of folds to train and evaluate over. In the following, the use of SDCC to construct training sets is investigated.

Consider a setting where CNNs are trained on images with a rotation angle of 15 degrees. Two CNNs are trained: (1) CNN$_{15}$, a CNN trained on images with a rotation angle of 15 degrees and (2) CNN$_{0, 15, 30}$, a CNN trained on images with rotation angles of 0, 15, and 30 degrees. Whereas CNN$_{15}$ is trained on 9k images with the same rotation, CNN$_{0, 15, 30}$ is trained on 4k images from three different angles, for a total of 12k images. The idea is that CNN$_{0, 15, 30}$ samples from nearby rotation angles during training for the sake of having a better coverage property.

CNN$_{15}$ and CNN$_{0, 15, 30}$ are trained for 20 epochs. Both classifiers are achieve an accuracy of 99\% on 1k held-out images with a rotation angle of 15 degrees. However, this does not hold when the distribution changes. Specifically, consider a case where CNN$_{15}$ and CNN$_{0, 15, 30}$ used to classify images with rotation angles 30, 45, and 60. Both models are tested on 10k images from each new angle.

The SDCC(\{30, 45, 60\}, 15) and SDCC(15, \{30, 45, 60\}) are 0.078 and 0.053, respectively. The SDCC(\{30, 45, 60\}, \{0, 15, 30\}) and SDCC(\{0, 15, 30\}, \{30, 45, 60\}) are slightly lower at 0.070 and 0.048, respectively. However, recall that SDCC is based on the presence or absence of value-combinations, where each image has 6 combinations of factors. While the SDCC values may be similar between the rotation angle 15 training set and the rotation angle 0, 15, and 30 training set, the number of images of rotation angle 30, 45, or 60 with value-combinations that are not covered varies greatly. 

Table \ref{table:train} shows that the CNN$_{15}$ training set has nearly 6-fold more uncovered images than the CNN$_{0,15,30}$ training set. While the classification accuracy of CNN$_{0,15,30}$ on rotation angles 30, 45, and 60 is slightly lower than CNN$_{15}$ for uncovered images, that lower accuracy applies to a much smaller fraction of the test set. Additionally, CNN$_{0,15,30}$ outperforms on covered images. So, other than the slightly lower accuracy on the far fewer uncovered images, CNN$_{0,15,30}$ dominates CNN$_{15}$ with higher accuracy on images with rotation angles 30, 45, and 60, lower SDCC scores for rotation angles 30, 45, and 60, and a lower number of uncovered images. All while having equal accuracy on images with rotation angle 15. This suggests that constructing training sets to have certain coverage-based properties has favorable trade-offs.

\subsection{Directing Labeling Efforts}

Based on the results presented thus far, it seems SDCC has a role to play in directing labeling efforts, e.g., as a metric in active or optimal learning \cite{settles2009active, powell2012optimal}. In other words, it seems that, given a set of unlabeled data, SDCC can help identify where algorithms will benefit the most from labeling efforts. Transfer learning from a source domain of non-rotated digits to target domains of various rotation angles is used to investigate this use of SDCC. Specifically, we consider the classification accuracy of CNN$_0$ on images with rotation angles of 60, 75, and 90 after fine-tuning CNN$_0$ with samples from corresponding angles.

The fine-tuning process follows a conventional transfer learning procedure of (1) initializing a new CNN$_\theta$ using the weights of CNN$_0$ and (2) training CNN$_\theta$ with images with a rotation angle of $\theta$, i.e., fine-tuning CNN$_0$ to a new angle $\theta$. For angle 60, 75, and 90, when the samples used in fine-tuning are drawn exclusively from images with not covered value combinations, classifier performance does not improve from the results reported in Table \ref{table:corr}. It may be that images with not covered value combinations represent edge cases, and that training on edge cases alone does not improve average performance.

\begin{figure*}[t]
    \centering
    \includegraphics[width=\textwidth]{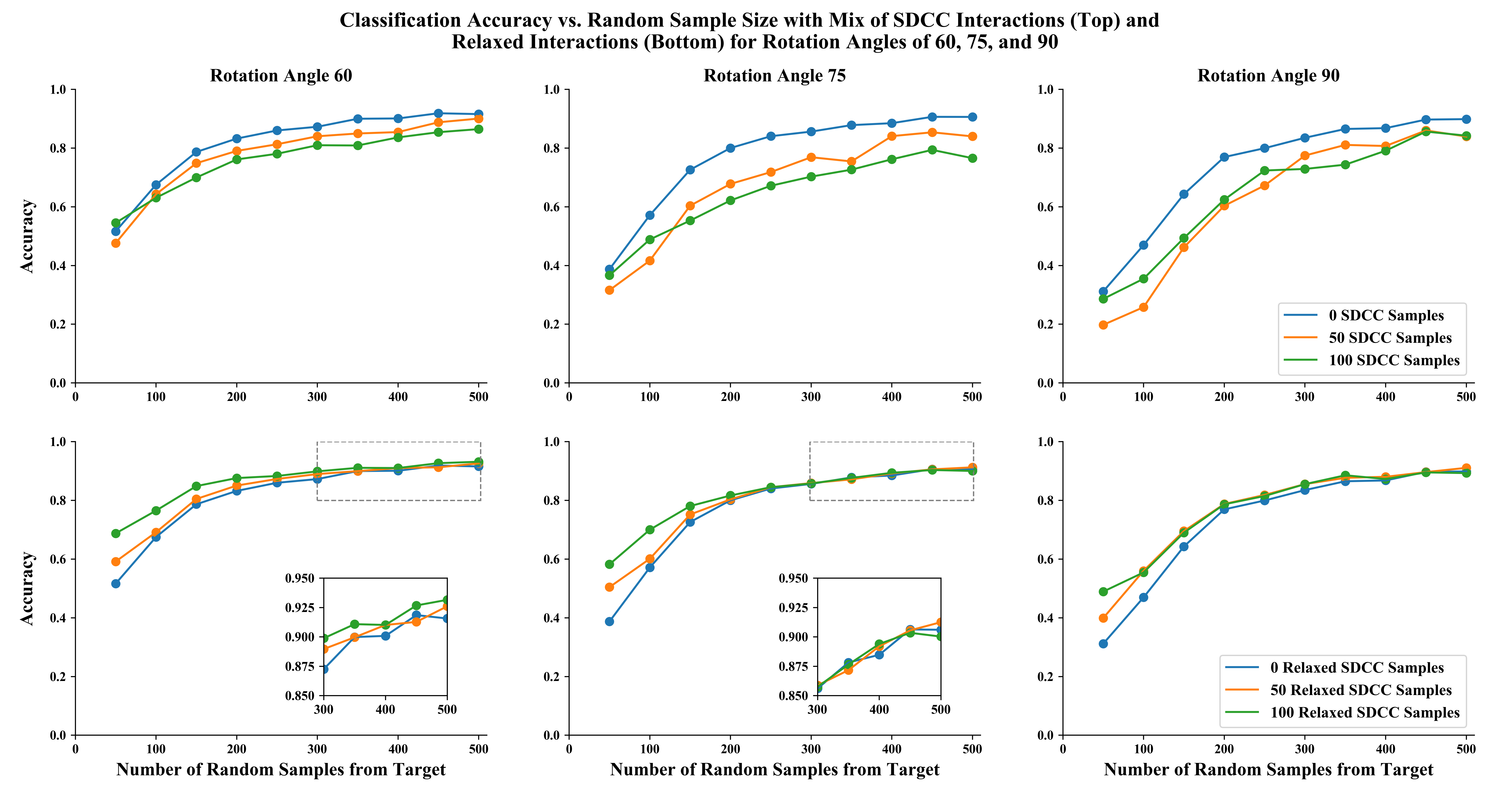}
    \caption{Impact of directed labeling on classification accuracy.}
    \label{fig:directed}
\end{figure*}

Alternatively, not covered images can be mixed into a random sample. The top row of Figure \ref{fig:directed} shows the result of this alternative. The y-axis denotes classification accuracy on 1k held-out images and the x-axis denotes the size of the random sample. The lines represent cases where 0, 50, and 100 not covered images are mixed with the random sample. By observing the total ordering of each case, it is clear that mixing not covered images into the random sample decreases sample efficiency and reduces the maximum accuracy.

\begin{table}[t]\centering
\ra{1.3}
\begin{tabular}{@{}rrrcrr@{}}\toprule
\multirow{2}{*}[-4pt]{{Train Set}}
& \multicolumn{2}{c}{Covered Images} & \phantom{abc}& \multicolumn{2}{c}{Not Covered Images}\\ 
\cmidrule{2-3} \cmidrule{5-6}
& \# Images & Accuracy && \# Images & Accuracy\\ \midrule
CNN$_{15}$ & 28663 & 0.75 && 1337 & 0.30\\
CNN$_{0, 15, 30}$ & 29768 & 0.85 && 232 & 0.23\\ \bottomrule
\end{tabular}
\caption{Accuracy of CNN$_{15}$ and CNN$_{0, 15, 30}$ on samples of images that are not covered by their respective training sets, as well as corresponding the corresponding number of images in those samples.}
\label{table:train}
\end{table}

The bottom row of Figure \ref{fig:directed} shows the results when the relaxed definition of covered is used in an otherwise identical experiment. Now, random samples mixed with SDCC samples outperform. They have a higher initial performance and they appear to converge faster. The left and middle plot of the bottom row show that this trend holds for larger random sample sizes. This suggests that while strict notions of SDCC can be used to construct difficult test sets and to construct robust training sets, the same capability of SDCC to select edge cases for training and testing may limit its use in directing labeling. However, using SDCC to guide which regions of the data space to sample from, as opposed to which exact pieces of data to sample, addresses this short-coming. It appears as if adding strictly not covered samples attempts to stretch the decision boundary over a disjoint set of points which degrades the global decision boundary. However, including weakly not covered samples helps maintain connectedness, so the decision boundary doesn't break.

A lack of utility from sampling of edge cases is highly dependent on the informativeness of factors, and could perhaps be assuaged with better factors, e.g., by using additional factors that are not invariant to rotation. Moreover, this short-coming may be a symptom of assessing average performance, rather than performance on edge cases.

\section{Rebuttal of Criticisms}

Coverage metrics have recently been criticized by Li et al. \cite{li2019structural}. Li et al. argue that the vast majority of the set of all possible value-combinations, i.e., the universal set $U$, consists of adversarial examples. Otherwise put, a vast minority of $U$ consists of so-called natural inputs. In the words of Li et al., ``This observation intuitively contradicts the effectiveness of the structural coverage criteria for neural networks.'' 

While this criticism may be true for DeepGauge \cite{ma2018deepgauge} or DeepCT \cite{ma2019deepct}, here, we demonstrate the richness of coverage by adjusting our concern away from notions of coverage relative to the universal set of value combinations $U$ to notions of set differences between the value-combinations in two different domains---that is, we use SDCC \cite{lanus2021combinatorial}. SDCC is a relative comparison between the presence of value-combinations in natural inputs, and therefore, is not based on comparisons to artificial, adversarial value-combinations.

Li et al. continue by claiming that, ``Our initial experiments with natural inputs denied the correlation between the number of misclassified inputs in a test set and its structural coverage on the associated neural networks.'' To the extent that natural inputs correspond to non-adversarial value-combinations in the set $U$, the results presented herein clearly falsify the initial experiments of Li et al., as shown explicitly in Table \ref{table:corr}.

Although Li et al. focus their criticisms on coverage methods which consider the neurons or layers of neural networks, the results presented herein rebut their criticisms by using methods which consider the data directly. Since this is equivalent to considering the input-layer of a neural network, this is a valid rebuttal of the criticisms quoted above. 

\section{Conclusion}

This paper presents adaptions and new methods for using combinatorial interaction testing in ML. The results show that SDCC correlates to test error, that model performance varies greatly between test sets with covered and not covered value combinations, that training sets constructed to satisfy coverage properties are more robust than identically distributed training sets, and that relaxed notions of coverage can improve label and sample efficiency. With these extensions, combinatorial interaction testing treats pressing issues in the training and testing in ML.

This and past works have used MNIST extensively, and, given the promising results, future work should extend the use of combinatorial interaction testing in machine learning to larger data sets and data sets with extensive meta-data. Also, while neuron- and layer-level notions of coverage may be powerful white-box testing methods, as demonstrated here, input-level notions of coverage can be powerful black-box testing methods. The use of combinatorial interaction testing as a spanning paradigm for both white- and black-box testing should be explored further.

\section*{Appendix}

\subsection{Autoencoder Architecture}

The AE takes flattened 28$\times$28 pixel images as input, has a single layer to encode to a dimension of 32 with a rectified linear unit activation function, and a single layer to decode back to a dimension of 784 with a sigmoid activation function. The model is trained using a batch size of 256, binary cross-entropy and Adam \cite{kingma2015adam}.

\subsection{CNN Architecture}

The CNNs used take 28$\times$28 pixel images as input, have hidden layers consisting of a 2-dimensional convolution layer with 32 filters of kernel size 3$\times$3, a layer of max pooling of size 2$\times$2, a 2-dimensional convolution layer with 64 filters of kernel size 3$\times$3, and another layer of 2$\times$2 max pooling. Dropout with a rate of 0.5 is used before the 10-dimensional output layer. The convolution layers use rectified linear units as activation functions and the output layer uses softmax. The model is trained using categorical cross-entropy and Adam \cite{kingma2015adam}. In training, the batch size was 128, except in the directing labeling study, which had smaller samples and used a batch size of 10.

\section*{Acknowledgements}

This research was sponsored by the Army Research Laboratory and was accomplished under Cooperative Agreement Number W911NF-18-2-0218. The views and conclusions contained in this document are those of the authors and should not be interpreted as representing the official policies, either expressed or implied, of the Army Research Laboratory or the U.S. Government. The U.S. Government is authorized to reproduce and distribute reprints for Government purposes notwithstanding any copyright notation herein.

\bibliographystyle{IEEEtran}
\bibliography{ref}

\end{document}